\documentclass[10pt,conference,a4paper]{IEEEtran}
\IEEEoverridecommandlockouts
\usepackage{times}
\usepackage{soul}
\usepackage{url}
\usepackage[utf8]{inputenc}
\usepackage{dashrule}
\usepackage{amsfonts}
\usepackage{graphicx}
\usepackage{amsmath}
\usepackage{amsthm}
\usepackage{booktabs}
\usepackage{bbding} %
\usepackage[ruled, vlined]{algorithm2e}
\usepackage{subfigure}
\usepackage{array}
\usepackage{longtable}
\usepackage{bm}
\usepackage{xcolor}
\usepackage{algpseudocode}

\newcommand{\ldotfill}[2]{\leavevmode\xleaders\hbox{\rule{2pt}{0.4pt}\ }\hfill\null}
\usepackage{graphicx}

\usepackage{multirow}

\usepackage[textwidth=3.6cm, textsize=footnotesize]{todonotes}

\usepackage{subfigure}
\usepackage{fmtcount}

\usepackage{amssymb}

\ifCLASSINFOpdf
\else
\fi
\begin{document}
	
	
%
\title{Softer Pruning, Incremental Regularization
}




\author{\IEEEauthorblockN{Linhang Cai\IEEEauthorrefmark{1}\IEEEauthorrefmark{2}, Zhulin An\IEEEauthorrefmark{1}${}^{\ddagger}$\thanks{${}^{\ddagger}$Zhulin An is the corresponding author.}, Chuanguang Yang\IEEEauthorrefmark{1}\IEEEauthorrefmark{2} and Yongjun Xu\IEEEauthorrefmark{1} }
	\IEEEauthorblockA{\IEEEauthorrefmark{1}Institute of Computing Technology, Chinese Academy of Sciences, Beijing, China}
	\IEEEauthorblockA{\IEEEauthorrefmark{2}University of Chinese Academy of Sciences, Beijing, China \\Email: \{cailinhang19g, anzhulin, yangchuanguang, xyj\}@ict.ac.cn}

}

\maketitle

\begin{abstract}
Network pruning is widely used to compress Deep Neural Networks (DNNs). The Soft Filter Pruning (SFP) method zeroizes the pruned filters during training while updating them in the next training epoch. Thus the trained information of the pruned filters is completely dropped. To utilize the trained pruned filters, we proposed a SofteR Filter Pruning (SRFP) method and its variant, Asymptotic SofteR Filter Pruning (ASRFP), simply decaying the pruned weights with a monotonic decreasing parameter. Our methods perform well across various networks, datasets and pruning rates, also transferable to weight pruning. 
On ILSVRC-2012, ASRFP prunes 40\% of the parameters on ResNet-34 with 1.63\% top-1 and 0.68\% top-5 accuracy improvement. In theory, SRFP and ASRFP are an incremental 
regularization of the pruned filters. Besides, We note that SRFP and ASRFP pursue better results while slowing down the speed of convergence.
\end{abstract}


%
\IEEEpeerreviewmaketitle

\section{Introduction}





Currently, deep Convolutional Neural Networks (CNNs) have shown extraordinary performance in various tasks, e.g., image classification~\cite{He2016,HCGNet}, target detection~\cite{Girshick2014}, semantic segmentation~\cite{Zhang201801}. However, large burdens of DNN model size, limited run-time memory and huge numbers of Floating Point Operations (FLOPs)~\cite{Liu2017} hinder the deployment of DNN models in mobile devices.
Thus, it matters to compress the DNN models to improve the computational efficiency. Prevalent methods of model compression and acceleration include low rank approximation~\cite{Jaderberg2014} and network pruning~\cite{Liu2018}. Among them, filter pruning can reduce both the model size and the computational cost, thus, having been a hot research topic.

\begin{figure}[t]
	\center
	\subfigure[Hard Filter Pruning]{
		\label{fig:hard_prune}
		\includegraphics[width=0.46\linewidth]{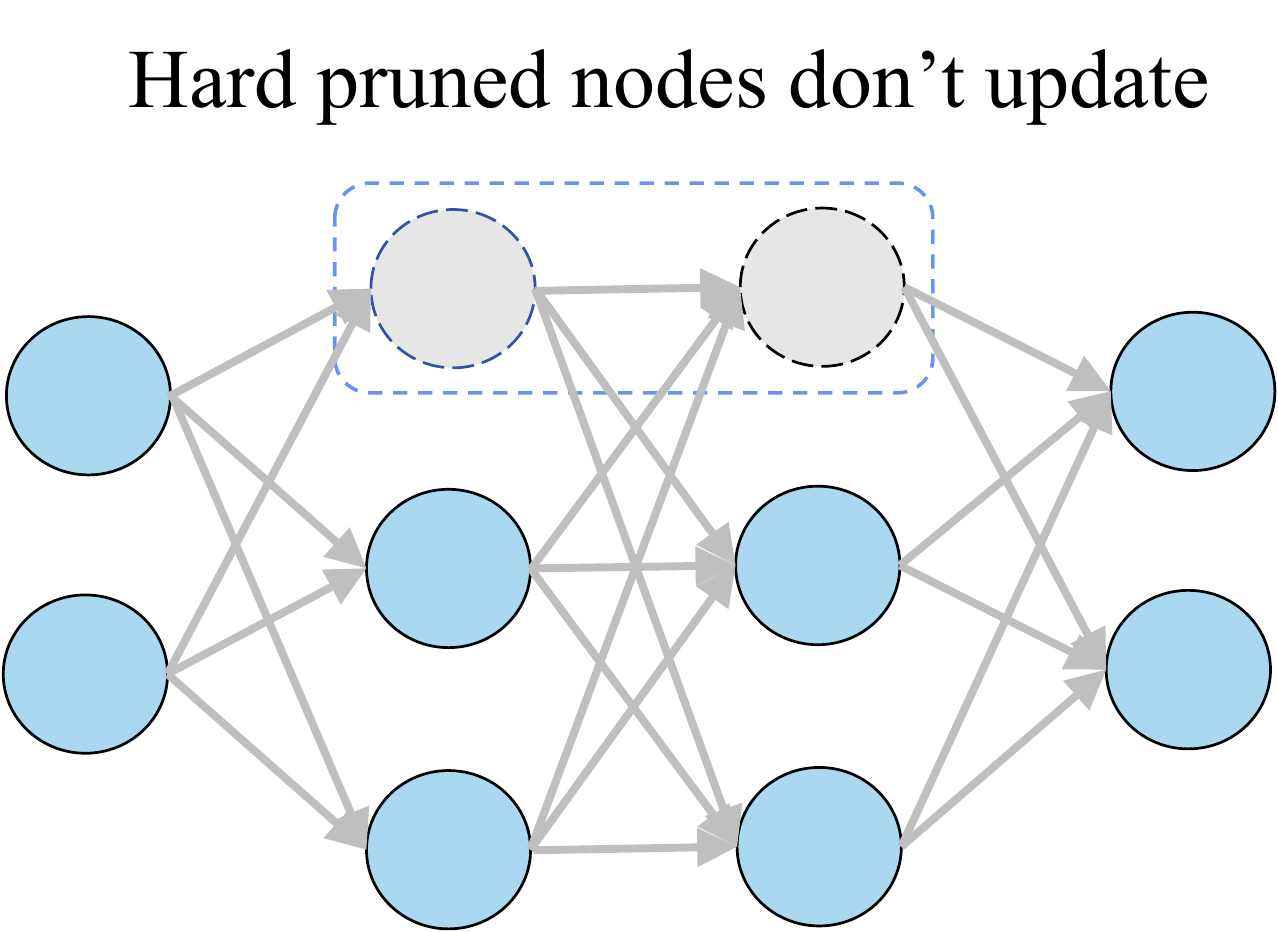}
	}
	\hspace{-2mm}
	\subfigure[Soft Filter Pruning]{
		\label{fig:soft_prune}
		\includegraphics[width=0.46\linewidth]{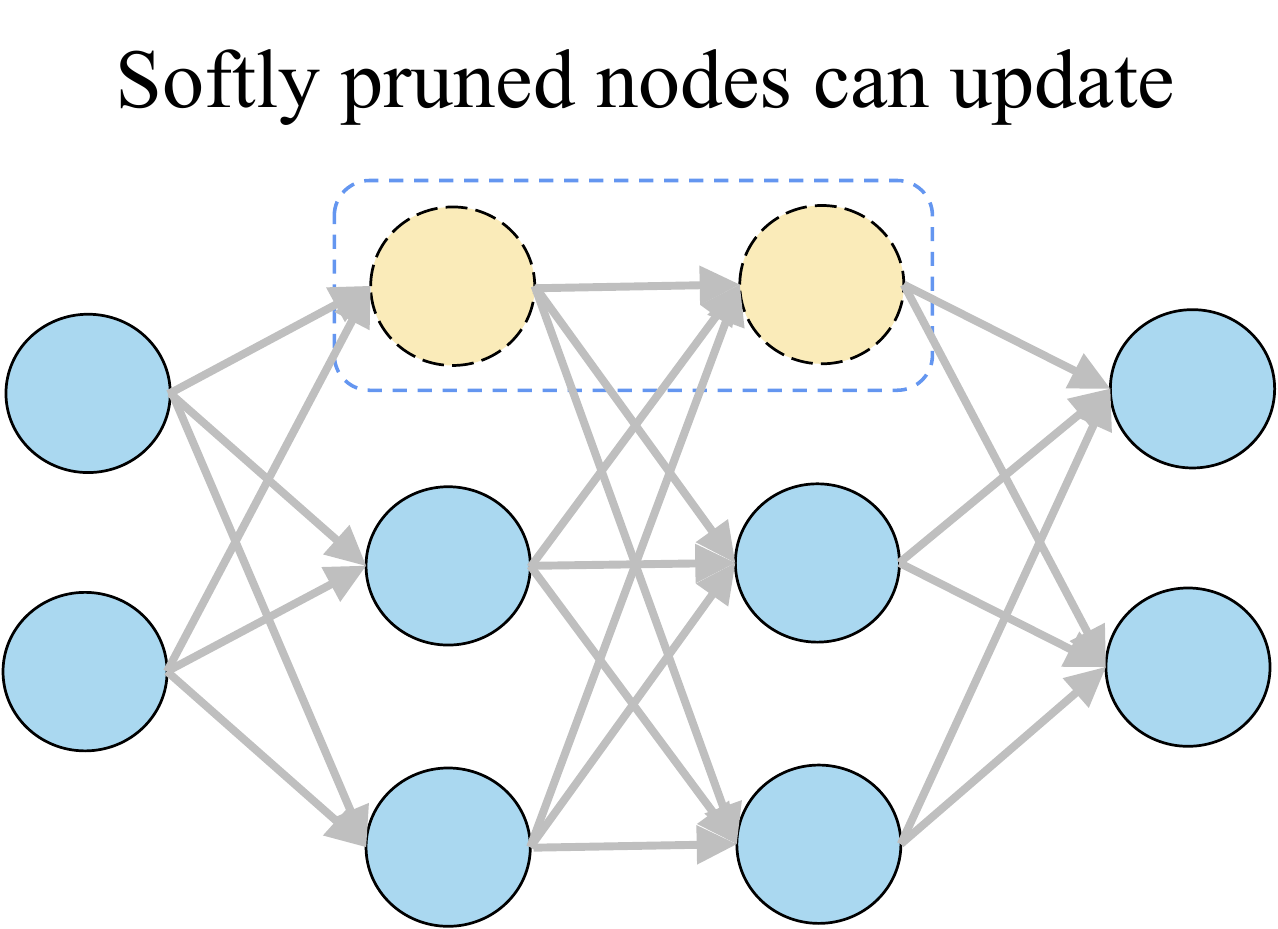}
	}
	\caption{
		Comparison of hard filter pruning and soft filter pruning. The gray nodes removed by hard filter pruning could not update in the next training epoch, while the yellow nodes pruned by soft filter pruning method can still update.
	}
	\vspace{-2.mm}
	\label{fig:hard_soft_prune}
\end{figure}

In terms of filter pruning, Soft Filter Pruning (SFP) maintains the capacity of the DNNs while pruning~\cite{He2018}. SFP zeroizes the filters chosen to be pruned and updates them in the next training epoch, as shown in Figure~\ref{fig:hard_soft_prune}. However, traditional hard filter pruning (HFP) method would not use those pruned filters any more.

A drawback of SFP is that there is a severe accuracy drop after pruning in case of large pruning rates. Asymptotic Soft Filter Pruning (ASFP) is a variant of SFP to stabilize the training and pruning process~\cite{He2019}. ASFP gradually increases the pruning rate towards the objective pruning rate to reduce the information loss caused by setting pruned filters to zeros while pruning.

\begin{figure}[!t]
	\vspace{-2.5mm}
	\begin{centering}
		\includegraphics[width=0.46\textwidth]{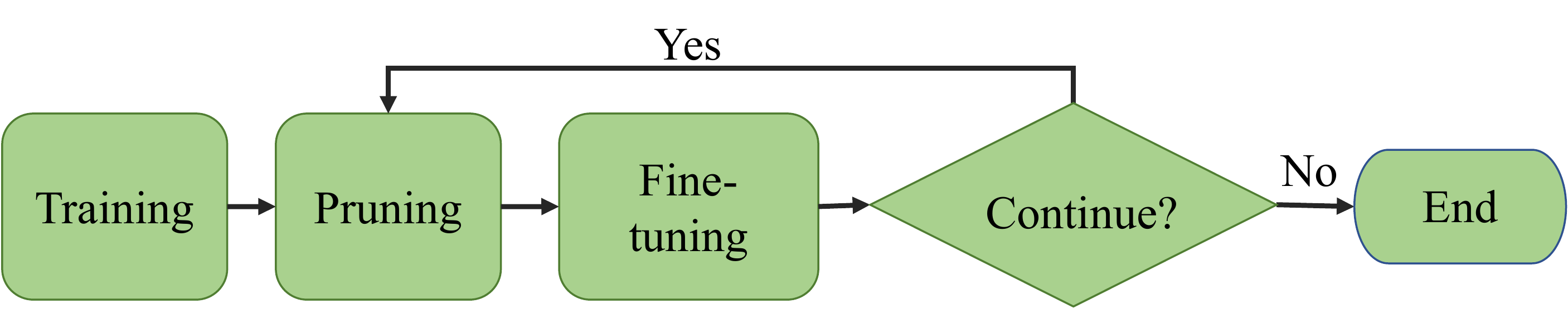} 
		\par\end{centering}
	\vspace{-1.mm}
	\caption{A typical three-step pruning pipeline composed of three phases: training, pruning and fine-tuning. Our SRFP or ASRFP is used in the pruning phase to remove those filters chosen to be pruned smoothly using weights that gradually decay to zero, while the conventional pruning operation simply sets pruned filters to zeros.
	}
	\vspace{-2mm}
	\label{fig:pruning_pipeline} 
\end{figure}

To utilize the trained pruned filters, we proposed a SofteR Filter Pruning (SRFP) method and its asymptotic version, Asymptotic SofteR Filter Pruning (ASRFP), simply decaying the pruned weights with a monotonic decreasing parameter. SRFP is indeed an incremental $\ell_2$-norm regularization of the pruned filters.

A typical three-step pruning pipeline consists of three phases: training, pruning and fine-tuning, as shown in Figure~\ref{fig:pruning_pipeline}. Our SRFP or ASRFP is used in the pruning phase to remove those filters chosen to be pruned smoothly using weights that gradually decay to zero. 

While the conventional pruning operation is to simply set parameters chosen to be pruned to zeros, our method ensures that the pruned filters are removed smoothly using weights that gradually decay to zero, so that we can better preserve the trained information in those filters. Our method, based on SFP, only needs tiny extra computations and hyperparameters, so it is easy and effective to use our method to prune a model.

With the same goal of finding minimal nets, our method is very different from the weight decay regularization~\cite{Loshchilov2019}. Actually, they are used in different phases. While weight decay is used in the training or fine-tuning phase to increase the sparsity of the networks or to avoid over-fitting, our method is used in the pruning phase to punish those pruned parameters in a soft manner. In short, while weight decay is used to constrain all weights in a network in the training or fine-tuning phase, our method only constrains those weights chosen to be pruned in the pruning phase. Our method will not affect those parameters that are not chosen to be pruned. In fact, in all of our experiments, weight decay is widely used in training and fine-tuning to avoid over-fitting.

Our contributions are as follows: (1) Gradually decaying the importance of the pruned filters, our methods perform well across various networks, datasets and pruning rates, also transferable to weight pruning. (2) The gradually decaying manner of SRFP and ASRFP is actually doing incremental regularization on those pruned weights, reserving the pruned weights at start, gradually forcing them towards zeros. (3) We 
note that SRFP, ASRFP and ASFP pursue better results while slowing down the speed of convergence.

\section{Related Works}

Previous attempts on deep neural network compression mainly include \emph{low rank matrix factorization}, \emph{fast convolution} and \emph{ pruning}.

Among them, \emph{low rank matrix factorization} is a tensor low rank expansion technique to reduce the number of parameters or speedup deep neural networks~\cite{Jaderberg2014}. However, these methods reveal  relatively small speedups on small convolutional kernels such as 3 $\times$ 3 and 1 $\times$ 1, which are widely used in prevalent CNNs like ResNets with bottleneck structures~\cite{He2016}. Compression-aware training~\cite{Alvarez2017} uses a regularizer to push the parameter matrix of each layer to be a low rank matrix as close as possible.

\emph{Fast Convolution} finds various efficient convolutional filters to design the efficient architecture, including Groupwise Convolution~\cite{Krizhevsky2012}, Depthwise Separable Convolution~\cite{Howard2017}, and Heterogeneous Kernel-Based Convolution (HetConv)~\cite{Singh2019}.

\emph{Pruning} techniques focus on reducing the network complexity by removing unimportant neural nodes or connections~\cite{Liu2018,Frankle2018}. Many studies on network pruning calculate the importance of the filters or connections and prunes them based on some criteria, and then finetune the pruned network to avoid severe accuracy drop. Some studies explore the automated determination of the threshold values for pruning~\cite{Manessi2018}.

\textbf{Weight Pruning.}
Magnitude-based weight pruning methods are computationally efficient, compressing networks by deleting unimportant weights in an unstructured manner. Iterative weight pruning is a three-step method to discard small weights whose magnitude below the threshold~\cite{Han201502}.
Dynamic network surgery properly incorporates both the pruning and splicing operations for model compression to avoid incorrect pruning, instead of alternately pruning and retraining~\cite{Guo2016}.


\textbf{Filter Pruning.}
Filter pruning removes redundant filters (channels) or feature maps. 
Prevalent metrics to evaluate the filter importance include $\ell_{1}$-norm, $\ell_{2}$-norm, scaling factors and feature redundancy~\cite{Li2016,Liu2017,Ayinde2018}. Gate Decorator~\cite{You2019} multiplies the output of a CNN module by channel-wise scaling factors, using Taylor expansion to evaluate the impact on the loss function owing to setting the scaling factor of each filter to zero. A common drawback of
these pruning techniques is that the network capacity is decreased after pruning. 

AutoPruner~\cite{Luo2020} combines the pruning phase and the fine-tuning phase of the typical three-step pruning pipeline into one single end-to-end trainable system to find unimportant filters automatically during training, thus increasing the computational and memory consumptions.

SFP and ASFP method set the pruned filters to zeros during training while updating them in the next training epoch to maintain the representative capacity.

During the first several pruning epochs of SFP, there is an obvious accuracy loss in the test set. ASFP gradually increases the pruning rate towards the aimed pruning rate to remedy this issue. To better utilize the trained pruned filters, we proposed a SofteR Filter Pruning (SRFP) method and its asymptotic version, Asymptotic SofteR Filter Pruning (ASRFP), simply decaying the pruned weights with a monotonic decreasing parameter $\alpha$. SRFP is an incremental regularization of the pruned filters. Our approach works well without much computational and memory consumptions. 

\section{Our Method}

\subsection{Formulation}\label{Preliminary}
Consider the convolutional kernel  ${W}_{i}\in\mathbb{R}^{n\times m\times s\times s}$ in the $i$-th layer, where $1\leq i\leq L$ and $L$ is the number of convolutional layers. Specifically, $s$, $m$ and $n$ are the convolutional kernel size, the number of input channels and output channels respectively.

The input feature map ${I}_i \in \mathbb{R}^{m \times h_{i}\times w_{i}} $ and the output feature map ${O}_i \in \mathbb{R}^{n \times h_{i+1}\times w_{i+1}}={W}_{i}\ast{I}_i$ is calculated by

\vspace{-3mm}
{\small
	\begin{align}
	\label{eq:1}
	{O}_{i,j}={W}_{i,j}\ast{I}_i ~&~ {for}~1 \leq j \leq n,
	\end{align}
}
\vspace{-3mm}

\noindent where ${O}_{i,j}  \in \mathbb{R}^{h_{i+1}\times w_{i+1}}$ and ${W}_{i,j} \in \mathbb{R}^{m\times s\times s}$ denote the $j$-th output channel and the $j$-th filter of the $i$-th layer respectively.

Suppose that the filter pruning rate for the $i$-th layer is $P_{i}$.
Thus there are $n\times P_{i}$ filters to be removed in the $i$-th layer,
and the size of the pruned output feature map ${O}_{i}$ would be ${n\times (1-P_{i})\times h_{i+1} \times w_{i+1}}$.

According to the SFP method and its variant ASFP, the pruned weights of the $i$-th layer are simply zeroized, which can be represented by

\vspace{-3mm}
{\small
	\begin{align}
	\label{eq:2}
	{\hat{W}}_{i,j}={W}_{i,j}\odot {M}_{i,j} ~&~ {for}~1 \leq j \leq n,
	\end{align}
}
\vspace{-3mm}

\noindent where ${M}_{i,j}$ is a Boolean matrix with the same shape as the filter ${W}_{i,j}$  to denote whether the $j$-th filter is pruned or not in the $i$-th layer. Besides, $\odot$ is a matrix pointwise multiplication operator. Exactly, ${M}_{i,j}=0$ if ${W}_{i,j}$ is pruned. Otherwise, we let ${M}_{i,j}=1$ to denote that the filter ${W}_{i,j}$ is not pruned.

\subsection{Motivation}\label{Preliminary}
Rewrite \eqref{eq:2} equivalently as

\vspace{-3mm}
{\small
	\begin{align}
	\label{eq:3}
	{\hat{W}}_{i,j}\!=\!{W}_{i,j}\!\odot\!{M}_{i,j}\!+\!
	\alpha\;\! {W}_{i,j}\!\odot\!(1\!-\!{M}_{i,j}) ~&~ {for}~1 \leq j \leq n,
	\end{align}
}
\vspace{-3mm}

\noindent where $\alpha=0$ in SFP as well as ASFP, and $\alpha$ is the decaying rate for those pruned weights. We can replace $\alpha$ with a decreasing nonzero number to better utilize the trained information inside those pruned weights. In general, we set $\alpha \in [0,1]$. 

Notably, when $\alpha \in (0,1]$, the trained knowledge of the pruned filters is not completely dropped which would be helpful for releasing the accuracy drop caused by the pruning phase and achieving better results in the next training epoch.

However, when $\alpha > 0$, the pruned filters are not zeroized so that the resulted pruned model is not compact. So we limit $\alpha$ to gradually decay from the initial value $\alpha_0$  where $\alpha_0 \in[0,1]$ towards zero as the training and pruning procedure goes on. We consider two kinds of decaying strategies, which are exponential decay and linear decay respectively.

The exponential decay strategy can be written as 

\vspace{-3mm}
{\small
	\begin{align}
	\label{eq:4}
	\alpha_e(t)=\alpha_0e^{ \frac{-k t}{t\_max-1}}  ~&~ {for}~0 \leq t < t\_max,
	\end{align}
}
\vspace{-3mm}

\noindent where $t\_max$ is the maximal number of training epochs and $k$ is a coefficient to control the descent speed of $\alpha$. Then we introduce a constraint parameter $\epsilon$ to obtain the value of $k$, claiming that

\vspace{-3mm}
{\small
	\begin{align}
	\label{eq:5}
	\alpha_e(t\_max-1)=\alpha_0e^{-k}  = \epsilon,~&~
	\end{align}
}
\vspace{-3mm}

\noindent where $\epsilon$ is infinitely close to zero.
Thus $k=\ln{\frac{\alpha_0}{\epsilon}} $ and $\alpha_e(t)$ can be given by

\vspace{-3mm}
{\small
	\begin{align}
	\label{eq:6}
	\alpha_e(t) = \alpha_0(\frac{\alpha_0}{\epsilon})^{-\frac{t}{t\_max-1}} ~&~ {for}~0 \leq t < t\_max,
	\end{align}
}
\vspace{-3mm}

\noindent when $\alpha_e(t)$ is approaching zero, we just set $\alpha_e(t)=0$ to obtain a really compact model.
Similarly, the linear decay strategy can be given by

\vspace{-3mm}
{\small
	\begin{align}
	\label{eq:7}
	\alpha_l(t)=\alpha_0(1 - \frac{t}{t\_max-1})  ~&~ {for}~0 \leq t < t\_max,
	\end{align}
}
\vspace{-3mm}

\noindent where $\alpha_l(0)=\alpha_0$ and $\alpha_l(t\_max-1) =0$.

\subsection{SofteR Filter Pruning (SRFP)}

Based on the above motivation, we illustrate our SRFP method in Algorithm~\ref{alg:SRFP}, where we prune $P_i\times 100\%$ of the filters in the $i$-th convolutional layer according to the $\ell_{2}$-norm of all filters. For simplicity, we use the same pruning rate $P_i$ for each convolutional layer to get rid of complicated hyper-parameter search.

At the beginning of the training and pruning phase, the pruned filters are decayed in a soft manner, especially when $\alpha$ is close to $1$, which means that we nearly maintain all the trained information inside the pruned filters. Thereby, we greatly avoid the sharp accuracy drop caused by pruning, achieving a better performance. As the phase goes on, we gradually push $\alpha$ towards $0$, making the training and pruning phase close to that of SFP method, in order to obtain an actually compact model.

Moreover, we can view our SRFP method from the angle of incremental regularization. Denote the pruned filters as  

\vspace{-3mm}
{\small
	\begin{align}
	\label{eq:8}
	{WP}_{i,j}={W}_{i,j}\odot(1-{M}_{i,j})  ~&~ {for}~1 \leq j \leq n,
	\end{align}
}
\vspace{-3mm}

\noindent where $1-{M}_{i,j}$ is regarded as a mask to obtain those pruned filters of the $i$-th layer. Thus \eqref{eq:3} equivalent to

\vspace{-3mm}
{\small
	\begin{align}
	\label{eq:9}
	{\hat{W}}_{i,j}={W}_{i,j}\odot{M}_{i,j}+\alpha {WP}_{i,j}  ~&~ {for}~1 \leq j \leq n,
	\end{align}
}
\vspace{-3mm}

\noindent where ${W}_{i,j}\odot{M}_{i,j}$ is the completely maintained while the pruned portion ${WP}_{i,j}$ is decayed by $\alpha$. We can split \eqref{eq:9} into the following two steps: 

\vspace{-3mm}
{\small
	\begin{align}
	\label{eq:10}
	{\hat{WP}}_{i,j}=\alpha {WP}_{i,j}  ~&~ {for}~1 \leq j \leq n,
	\end{align}
}
\vspace{-3mm}

\noindent where ${\hat{WP}}_{i,j}$ is ${WP}_{i,j}$ decayed by $\alpha$, and then 

\vspace{-3mm}
{\small
	\begin{align}
	\label{eq:11}
	{\hat{W}}_{i,j}={W}_{i,j}\odot{M}_{i,j}+{\hat{WP}}_{i,j}  ~&~ {for}~1 \leq j \leq n,
	\end{align}
}
\vspace{-3mm}

\noindent where ${\hat{W}}_{i,j}$ is the initial value of the $j$-th filter in the $i$-th layer for the next training epoch.

Denote $\lambda(t) = \alpha_0-\alpha(t)$, where $\alpha(t)$ could be either exponential or linear decay. Thus we rewrite \eqref{eq:10} equivalently as

\vspace{-3mm}
{\small
	\begin{align}
	\label{eq:12}
	{\hat{WP}}_{i,j}=&\alpha {WP}_{i,j}=(\alpha_0-\lambda ) {WP}_{i,j}=\alpha_0 {WP}_{i,j} -\lambda {WP}_{i,j} \nonumber \\=& \alpha_0 {WP}_{i,j} -\frac{\lambda}{2}\frac{\partial \|{WP}_{i,j}\|_2^2}{\partial {WP}_{i,j}} ~~~ {for}~1 \leq j \leq n.
	\end{align}
}
\vspace{-3mm}

\noindent Consider the special case when $\alpha_0=1$. Our SRFP is indeed adding a $\ell_2$-norm regularization term $\frac{\lambda }{2}\|   {WP}_{i,j}\|_2^2$ to those pruned filters. Since $\alpha(t)$ decreases from $1$ to $0$, then $\lambda(t)$ increases from $0$ to $1$. So the regularization gradually strengthened.  

Besides, as the asymptotic variant of SFP called ASFP gradually increases the pruning rate towards the final pruning rate, likewise, we also create an asymptotic version of SRFP named ASRFP that gradually increases the pruning rate.

\begin{algorithm}
	\caption{SRFP Algorithm}\label{alg:SRFP}
	\footnotesize
	\SetKwInOut{Input}{inputs}\SetKwInOut{Output}{output}
	
	\Input{training set: ${X}$, pruning rate: $P_{i}$, initial decay rate: $\alpha$, \\the model with parameters ${W} = \{{W}_{i}, 0\leq i \leq L\}$. }
	\Output{The pruned model with parameters ${W} ^{*}={W} ^{t\_max}$}
	Initialize the model parameter ${W}^0$   \\
	
	\For{$t=0$, ..., $ t\_{max}-1$}{
		Decrease weight decay rate $\alpha$ \\
		Train model parameters $\hat{W}^{t+1}$ based on data set ${X}$ and  ${W}^{t}$\\
		\For{$i=1$, ..., $L$}{
			Compute the $\ell_2$-norm of each filter $\|\hat{W}_{i,j}^{t+1}\|_2, 1 \leq j \leq n$ \\
			Select $n\times P_i$ filters with minimal $\ell_2$-norm values\\
			Decay the parameters of chosen filters with $\alpha$\\
				
		}
		Get the softly pruned model parameters ${W}^{t+1}$ based on $\hat{W}^{t+1}$
		
	}
	Get the pruned model with final parameters ${W} ^{*}={W} ^{t\_max}$  \\
\end{algorithm}
\vspace{-2mm}

\begin{table*}[th]
	\setlength{\tabcolsep}{1.2mm}
	\caption{COMPARISON OF PRUNING RESNETS ON CIFAR-10
	} 	
	\vspace{-1mm}
	\label{table:cifar10_accuracy} 
	\small  	
	\centering  	
	\begin{tabular}{c|c c c c c c c}  		
		\toprule 
		Depth & Method & {Pre-trained?} &Baseline Accu. (\%)  &Accelerated Accu. (\%)  &Accu. Drop (\%) & FLOPs & Pruned FLOPs(\%)       
		\\ \midrule 
		\multirow{10}{*}{56} 		 &PFEC~\cite{Li2016} &
		$\times$& 93.04                   & 91.31  &1.75 &9.09E7  &27.6 \\
		&\vspace{-4mm} \ldotfill{4pt}{1pt} & \ldotfill{14pt}{11pt} & \ldotfill{14pt}{11pt} & \ldotfill{14pt}{11pt}&
		\ldotfill{14pt}{11pt}& \ldotfill{14pt}{11pt} & \ldotfill{14pt}{11pt}	 \\ 
		
		&CP~\cite{He2017} &$\times$& 92.80                      & 90.90 &1.90 &	-	  & 50.0\\   
		
		& SFP(20\%) &$\times$  &{93.66 $\pm$ 0.28}   & 93.26 $\pm$ 0.20 & 0.40 &8.98E7 &28.4\\              
		& ASFP(20\%)  &$\times$ &{93.66 $\pm$ 0.28}  & 93.26 $\pm$ 0.21 & 0.40   &  8.98E7 &28.4 \\              
		& Our1(20\%) &$\times$  &{93.66 $\pm$ 0.28}   & \textbf{93.33} $\pm$ 0.43 & \textbf{0.33} &8.98E7 &28.4\\ 
		&\vspace{-4mm} \ldotfill{4pt}{1pt} & \ldotfill{14pt}{11pt} & \ldotfill{14pt}{11pt} & \ldotfill{14pt}{11pt}&
		\ldotfill{14pt}{11pt}& \ldotfill{14pt}{11pt} & \ldotfill{14pt}{11pt}	 \\              
		& Our2(20\%) &$\times$  &{93.66 $\pm$ 0.28}  & 92.92 $\pm$ 0.51 & 0.74   & 8.98E7 &28.4 \\  
		& SFP(20\%) &\checkmark  &{93.34 }  & 93.25  & 0.09   & 8.98E7 &28.4 \\
		& ASFP(20\%) &\checkmark  &{93.34 }  & 93.23  & 0.11   & 8.98E7 &28.4 \\
		& Our1(20\%) &\checkmark  &{93.34 }  & 93.17  & 0.17   & 8.98E7 &28.4 \\ 
  
		& Our2(20\%) &\checkmark  &{93.34 }  & 93.25  & 0.09   & 8.98E7 &28.4\vspace{-3.0mm} \\   
		
						&\vspace{0.2mm} \ldotfill{4pt}{1pt} & \ldotfill{14pt}{11pt} & \ldotfill{14pt}{11pt} & \ldotfill{14pt}{11pt}&
		\ldotfill{14pt}{11pt}& \ldotfill{14pt}{11pt} & \ldotfill{14pt}{11pt}	 \\ 
		
		& SFP(40\%) &$\times$  &{93.66 $\pm$ 0.28}   & 92.06 $\pm$ 0.68 & 1.60 &5.94E7 &52.6\\              
		& ASFP(40\%) &$\times$  &{93.66 $\pm$ 0.28}  & 92.46 $\pm$ 0.43 & 1.20   & 5.94E7 &52.6 \\              
		& Our1(40\%) &$\times$  &{93.66 $\pm$ 0.28}   & {92.67} $\pm$ 0.45 & {0.99} &5.94E7 &52.6\\              
		& Our2(40\%)&$\times$   &{93.66 $\pm$ 0.28}  & \textbf{92.92} $\pm$ 0.39 & \textbf{0.74}   & 5.94E7 &52.6 \\       
		\midrule 
		
		\multirow{10}{*}{110}          		&PFEC~\cite{Li2016} &$\times$&93.53     & 92.94      &  0.61              & 1.55E8 	&38.6 	 \\       
		&\vspace{-4mm} \ldotfill{4pt}{1pt} & \ldotfill{14pt}{11pt} & \ldotfill{14pt}{11pt} & \ldotfill{14pt}{11pt}&
		\ldotfill{14pt}{11pt}& \ldotfill{14pt}{11pt} & \ldotfill{14pt}{11pt}	 \\   
		&MIL~\cite{Dong2017} &$\times$& 93.63         & 93.44 &0.19 &	-	  & 34.2 	 \\           		
		& SFP(20\%) &$\times$  & {94.33 $\pm$ 0.40}	&\textbf{93.86} $\pm$ 0.45 & \textbf{0.47 }&  1.82E8 &28.2 \\
		& ASFP(20\%) &$\times$  & {94.33 $\pm$ 0.40}	&{93.62 $\pm$ 0.53} & {0.71 }&  1.82E8 &28.2 \\  
		& Our1(20\%)  &$\times$ & {94.33 $\pm$ 0.40}	&{93.61 $\pm$ 0.56} & {0.72 }&  1.82E8 &28.2 \\ 
		&\vspace{-4mm} \ldotfill{4pt}{1pt} & \ldotfill{14pt}{11pt} & \ldotfill{14pt}{11pt} &
		\ldotfill{14pt}{11pt}& \ldotfill{14pt}{11pt}& \ldotfill{14pt}{11pt} & \ldotfill{14pt}{11pt}	 \\ 
		& Our2(20\%)&$\times$   & {94.33 $\pm$ 0.40}	&{93.83 $\pm$ 0.58} & {0.50 }&  1.82E8 &28.2 \\ 
		& SFP(40\%)  &$\times$ & {94.33 $\pm$ 0.40}	&{92.91} $\pm$ 0.53 & {1.42 }&  1.21E8 &52.3 \\
		& ASFP(40\%)&$\times$   & {94.33 $\pm$ 0.40}	&{93.52 $\pm$ 0.19} & {0.81 }& 1.21E8 &52.3 \\  
		& Our1(40\%)  &$\times$ & {94.33 $\pm$ 0.40}	&{93.66 $\pm$ 0.19} & {0.67 }& 1.21E8 &52.3 \\  
		& Our2(40\%)  &$\times$ & {94.33 $\pm$ 0.40}	&\textbf{93.69} $\pm$ 0.24 & \textbf{0.64 }&  1.21E8 &52.3 \\    		 \bottomrule 
	\end{tabular}
\end{table*}

\section{Experimental Results}\label{Experiment}

\subsection{Setup}

We evaluate our approaches on the CIFAR-10~\cite{Krizhevsky2009}, and ILSVRC-2012~\cite{Russakovsky2015}.
CIFAR-10 includes 60,000 RGB images of size $32 \times 32$ pixels, divided into 10 classes. ILSVRC-2012 consists of 1.28 million training
images and 50k validation images drawn from 1,000 categories. We focus on pruning the prevalently utilized ResNet, following the common experimental setup in ThiNet~\cite{Luo2017}, CP~\cite{He2017}.

On CIFAR-10, we follow the parameter scheme and the training configuration in~\cite{He201602}. On ILSVRC-2012, we follow the parameter scheme as~\cite{He201602},
and adopt the same data-augmentation
scheme as ~\cite{Paszke2017}.

Our SRFP or ASRFP is adopted after finishing a training epoch. Models are trained from scratch by default. We also provided results with pre-trained models, where the learning rate is one tenth of that of models trained from scratch. The experiments are repeated five times, reported by the ``mean $\pm$ std''. Then we present the results comparing with other state-of-the-art methods, e.g., SFP~\cite{He2018}, ASFP~\cite{He2019}, MIL~\cite{Dong2017}, PFEC~\cite{Li2016}, CP~\cite{He2017}, ThiNet~\cite{Luo2017}, AutoPruner~\cite{Luo2020}.

\subsection{ResNet on CIFAR-10}

\begin{figure}[!t]
	\vspace{-2.5mm}
	\begin{centering}
		\includegraphics[width=0.43\textwidth]{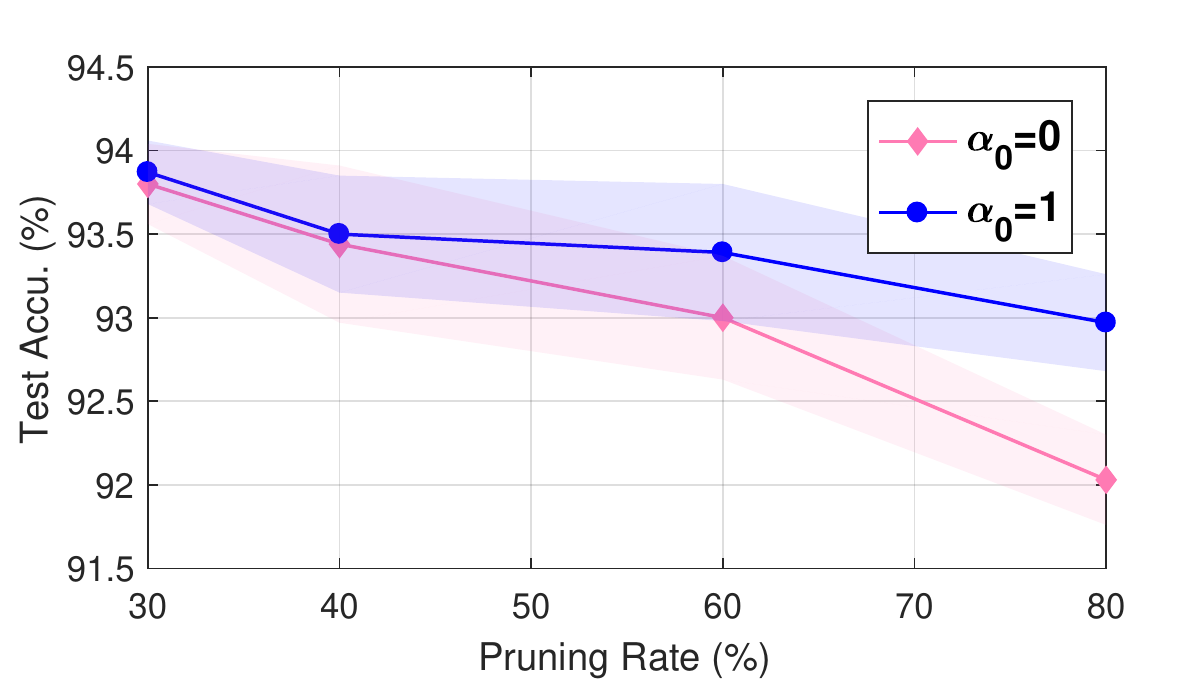} 
		\par\end{centering}
	\vspace{-1.mm}
	\caption{
		Transferring to Weight Pruning. Comparison of ResNet-56 on CIFAR-10 between SFP ($\alpha_0=0$) and our SRFP ($\alpha_0=1$) with the linear decay strategy. (Solid line and shadow represent the mean and standard deviation individually.)
	}
	\vspace{-6mm}
	\label{fig:wp} 
\end{figure}

\textbf{Settings}.  On CIFAR-10, we evaluate our SRFP and ASRFP on ResNet-20/56/110, adopting various pruning rates to study the efficacy of our methods, especially comparing with that of SFP and ASFP. By default, we use exponential decay with $\epsilon=1e-5$ and $\alpha_0=1$.

\begin{figure*}[t]
	\center
	\subfigure[ResNet-20]{
		\label{fig:sfp_asfp_cifar10_20}
		\includegraphics[width=0.28\linewidth]{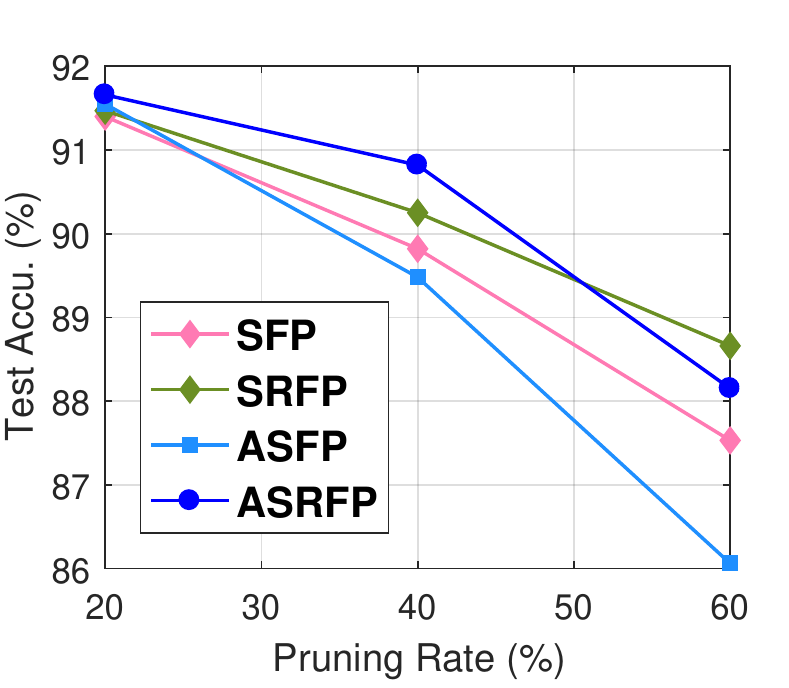}
	}
	\hspace{-6mm}
	\subfigure[ResNet-56]{
		\label{fig:sfp_asfp_cifar10_56}
		\includegraphics[width=0.28\linewidth]{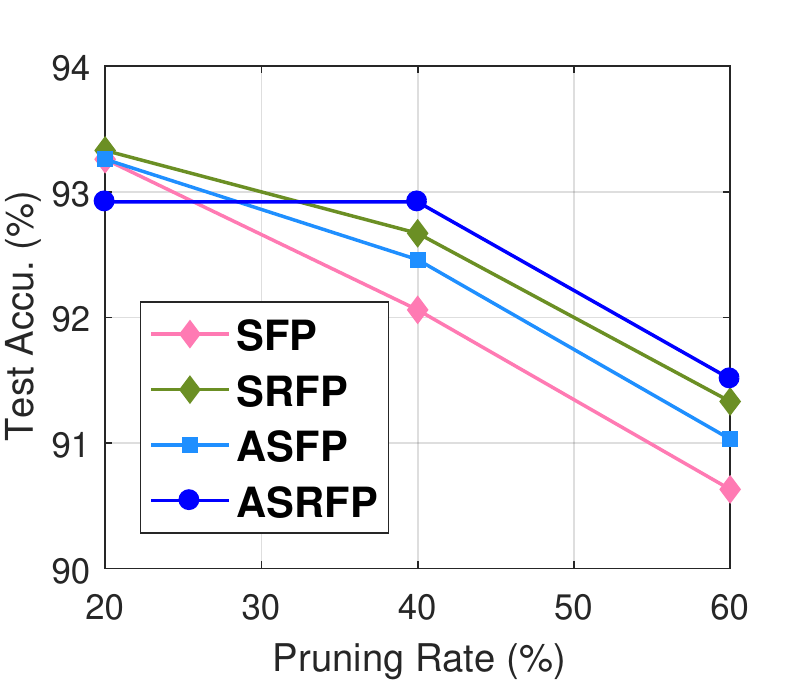}
	}
	\hspace{-6mm}
	\subfigure[ResNet-110]{
		\label{fig:sfp_asfp_cifar10_110}
		\includegraphics[width=0.28\linewidth]{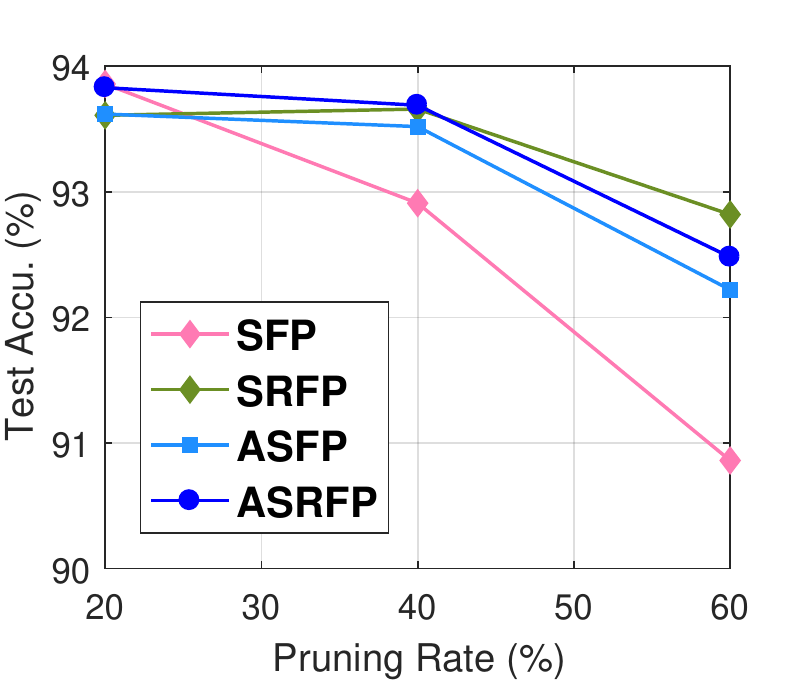}
	}
	\vspace{-2mm}
	\caption{
		Comparison of Test Accuracies of ResNet-20/56/110 on CIFAR-10 among SFP/ASFP/SRFP/ASRFP with the pruning rate changing.
	}
	\vspace{-0mm}
	\label{fig:sfp_asfp_cifar10}
\end{figure*}



\textbf{Results of Filter Pruning}. We conclude the results of both SRFP and ASRFP on CIFAR-10 
in Table~\ref{table:cifar10_accuracy}
. 
Here, we use "Our1" and "Our2" to refer to SRFP and ASRFP respectively for clarity. We mainly compare our methods with SFP and ASFP. Both SRFP and ASRFP reveal competitive performance on 
CIFAR-10 
compared with other channel pruning techniques across networks of various depths and pruning rates. Notably, our ASRFP performs better than other methods in most cases in Table~\ref{table:cifar10_accuracy}
. The models are trained from scratch, and the ``Accu.~Drop'' is the accuracy of the pruned model minus that of the baseline model.
The smaller is the better.

We compare the results of SFP and SRFP with various pruning rates and network depths across CIFAR-10, shown in Figure~\ref{fig:sfp_asfp_cifar10}
. When the pruning rate is as small as 20\%, the behaviors of the above four methods are quite similar. As the pruning rate increases to 60\%, our SRFP and ASRFP outperform SFP and ASFP. 

\begin{figure}[!t]
	\vspace{2.mm}
	\begin{centering}
		\includegraphics[width=0.52\textwidth]{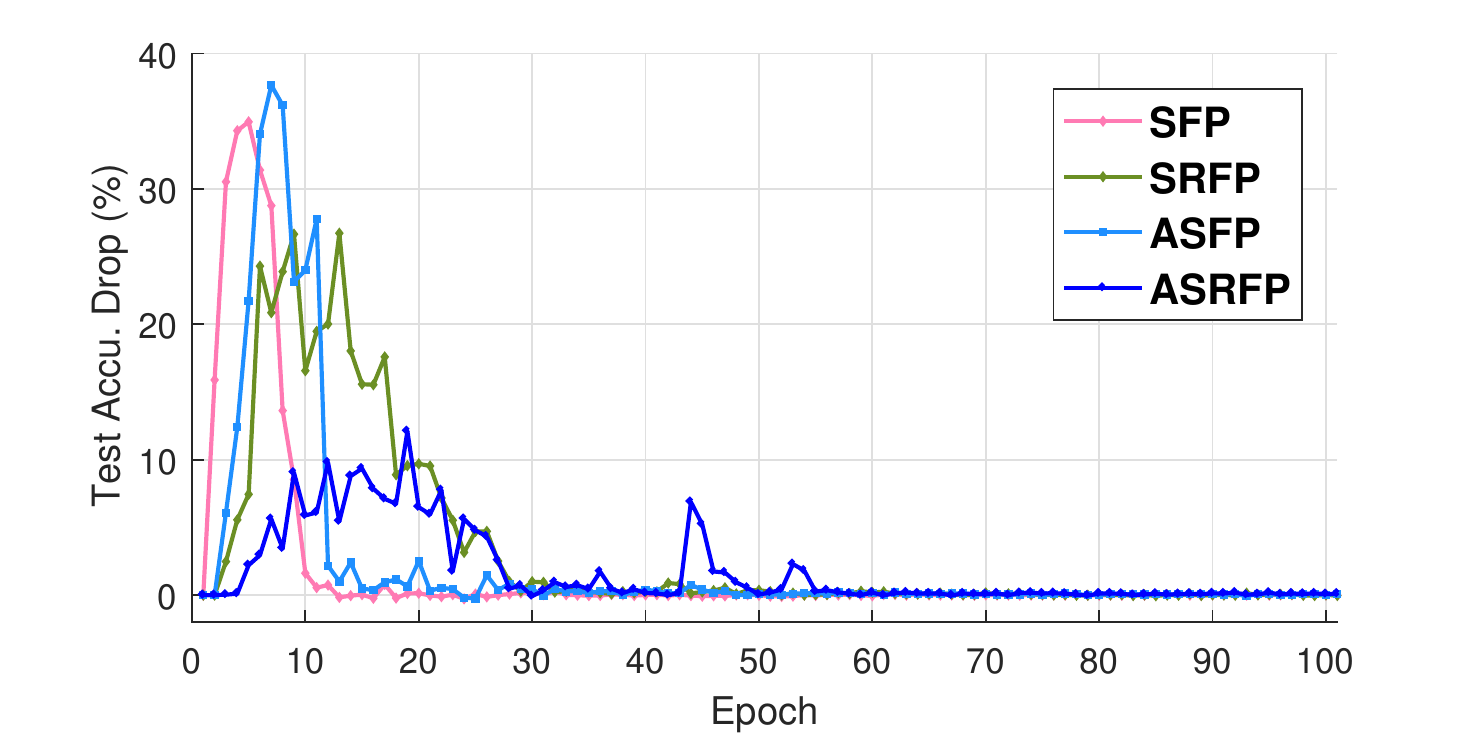} 
		\par\end{centering}
	\vspace{-1.mm}
	\caption{Comparison of different Test Accuracy Drops of ResNet-34 on ILSVRC-2012 among SFP/ASFP/SRFP/ASRFP with the training epochs increasing when the pruning rate is 30\%. The Test Accuracy Drop is the difference between the Top-1 accuracy before pruning and the Top-1 accuracy after pruning, where 0 means that there is no obvious accuracy drops caused by pruning.
	}
	\vspace{-2mm}
	\label{fig:acc_gap} 
\end{figure}

\textbf{Transferability to Weight Pruning}. Since filter pruning is a special case of weight pruning, we test the transferability of our SRFP method to weight pruning on CIFAR-10, pruning ResNet-56 with diverse pruning rates, using the linear decay strategy defined by \eqref{eq:7}. The results are shown in Figure~\ref{fig:wp}, from which we can verify the transferability of our ASFP to weight pruning. Notably, our ASFP has larger relative advantages than SFP in case of larger pruning rates.


\subsection{ResNet on ILSVRC-2012}
\label{section:ILSVRC}

\setlength{\tabcolsep}{0.9em} 

\setlength{\tabcolsep}{0.1em} 
\begin{table*}[ht] \small
	\caption{
		COMPARISON OF PRUNING RESNETS ON IMAGENET
	}  
	\vspace{-1mm} 	
	\label{table:imagenet_accuracy}
	\centering 	
	\begin{tabular}{c|c c c c c c c c c} 		
		\toprule 
		\multirow{2}{*}{Depth}	   & \multirow{2}{*}{Method}   &
		\multirow{2}{*}{Pre-trained?}
		 &\multirow{2}{*}{\shortstack {Top-1 Accu.\\Baseline(\%)} }  &\multirow{2}{*}{\shortstack {Top-1 Accu.\\Accelerated(\%)} } &\multirow{2}{*}{\shortstack {Top-5 Accu.\\Baseline(\%)} }   &\multirow{2}{*}{\shortstack {Top-5 Accu.\\Accelerated(\%)} }   &\multirow{2}{*}{\shortstack {Top-1 Accu.\\ Drop(\%)} }  &\multirow{2}{*}{\shortstack {Top-5 Accu.\\ Drop(\%)} } & \multirow{2}{*}{\shortstack {Pruned\\FLOPs(\%)}}    \\
		& & & & & & & &  \\ 
		\midrule 
		\multirow{6}{*}{18} 		
		& \vspace{-3.mm} & & & & & & & & \\
		&\vspace{-4mm} \ldotfill{4pt}{1pt} & \ldotfill{14pt}{11pt} & \ldotfill{14pt}{11pt} &
		\ldotfill{14pt}{11pt} &
		 \ldotfill{14pt}{11pt}& \ldotfill{14pt}{11pt} & \ldotfill{14pt}{11pt} \ldotfill{14pt}{11pt} & \ldotfill{14pt}{11pt} \ldotfill{14pt}{11pt} & \ldotfill{14pt}{11pt}	 \\  
		&MIL~\cite{Dong2017} & $\times$   &69.98 &66.33 & 89.24     & 86.94	&3.65	 &2.30   & 34.6		 \\	       		        
		& SFP(30\%)~\cite{He2018} & $\times$   & {70.23}	&{67.10} &{89.51}  & {87.78}   &{3.13}   &{1.73} & {41.8} \\
		& ASFP(30\%)~\cite{He2019} & $\times$  & {70.23}	&{67.26} &{89.51}  & {87.88}   &{2.97}   &{1.63} & {41.8} \\
		& Our1(30\%) & $\times$  & {70.23}	&\textbf{68.06} &{89.51}  & \textbf{88.06}   &\textbf{2.17}   &\textbf{1.45} & {41.8} \\
		&\vspace{-4mm} \ldotfill{4pt}{1pt} & \ldotfill{14pt}{11pt} & \ldotfill{14pt}{11pt} &
		\ldotfill{14pt}{11pt} & \ldotfill{14pt}{11pt}& \ldotfill{14pt}{11pt} & \ldotfill{14pt}{11pt} \ldotfill{14pt}{11pt} & \ldotfill{14pt}{11pt} \ldotfill{14pt}{11pt} & \ldotfill{14pt}{11pt}	 \\
		& Our2(30\%) & $\times$  & {70.23}	&{67.25} &{89.51}  & {87.59}   &{2.98}   &{1.92} & {41.8} \\
		& ASFP(40\%)~\cite{He2019} & $\times$  & {70.23}	&{65.44} &{89.51}  & {86.47}   &{4.79}   &{3.04} & {53.5} \\
		& Our2(40\%) & $\times$  & {70.23}	&\textbf{65.56} &{89.51}  & \textbf{86.58}   &\textbf{4.67}   &\textbf{2.93} & {53.5} \\
		\midrule 
		
		\multirow{7}{*}{34} 
		& \vspace{-3.mm} & & & & & & & \\
		&\vspace{-4mm} \ldotfill{4pt}{1pt} & \ldotfill{14pt}{11pt} & \ldotfill{14pt}{11pt} & \ldotfill{14pt}{11pt}&
		\ldotfill{14pt}{11pt} & \ldotfill{14pt}{11pt} & \ldotfill{14pt}{11pt} \ldotfill{14pt}{11pt} & \ldotfill{14pt}{11pt} \ldotfill{14pt}{11pt} & \ldotfill{14pt}{11pt}	 \\	          
		&MIL~\cite{Dong2017} & $\times$    & 73.42          &{72.99} 	&91.36	 &{91.19}	&{0.43} 	&{0.17} 	& 24.8	\\          
		& SFP(30\%)~\cite{He2018} & $\times$ 	 	&{73.92}		&71.15	&{91.62}   & 89.54   & 2.77  & 1.67 & {41.1}    \\
		& ASFP(30\%)~\cite{He2019}	& $\times$  	&{73.92}		&71.17	&{91.62}   & 90.11   & 2.75  & 1.46 & {41.1}    \\
		& Our1(30\%)	& $\times$  	&{73.92}		&71.35	&{91.62}   & 90.20   & 2.57  & 1.42 & {41.1}    \\
		&\vspace{-4mm} \ldotfill{4pt}{1pt} & \ldotfill{14pt}{11pt} & \ldotfill{14pt}{11pt} & \ldotfill{14pt}{11pt}& \ldotfill{14pt}{11pt} &
		\ldotfill{14pt}{11pt} & \ldotfill{14pt}{11pt} \ldotfill{14pt}{11pt} & \ldotfill{14pt}{11pt} \ldotfill{14pt}{11pt} & \ldotfill{14pt}{11pt}	 \\			  
		& Our2(30\%)	& $\times$  	&{73.92}		&\textbf{71.39}	&{91.62}   & \textbf{90.23}   & \textbf{2.53}  & \textbf{1.39} & {41.1}    \\    
		& ASFP(40\%)~\cite{He2019}	 & $\times$ 	&{73.92}		&68.79	&{91.62}   & 88.95   & 5.13  & 2.67 & {52.7}    \\		  
		& Our1(40\%)	 & $\times$ 	&{73.92}		&\textbf{70.73}	&{91.62}   & \textbf{89.87}   & \textbf{3.19}  & \textbf{1.75} & {52.7}    \\   
		& Our2(40\%)	 & $\times$ 	&{73.92}		&{70.42}	&{91.62}   & {89.63}   & {3.50}  & {1.99} & {52.7}    \\  
		\midrule 
		
		\multirow{4}{*}{50}
		& SFP(40\%)~\cite{He2018}  & $\times$ &{76.15}		&{73.04}		&{92.87}	  & {91.40} &3.11	 & 1.47 & 53.5  	 \\	     		  		  
		& ASFP(40\%)~\cite{He2019}  & $\times$ &{76.15}		&{72.98}		&{92.87}	  & {91.48} &3.17	 & 1.39 & 53.5  	 \\
		& Our1(40\%)  & $\times$ &{76.15}		&\textbf{73.62}		&{92.87}	  & \textbf{91.74} &\textbf{2.53}	 & \textbf{1.13} & 53.5  	 \\
		&\vspace{-4mm} \ldotfill{4pt}{1pt} & \ldotfill{14pt}{11pt} & \ldotfill{14pt}{11pt} & \ldotfill{14pt}{11pt}& \ldotfill{14pt}{11pt} &
		\ldotfill{14pt}{11pt} & \ldotfill{14pt}{11pt} \ldotfill{14pt}{11pt} & \ldotfill{14pt}{11pt} \ldotfill{14pt}{11pt} & \ldotfill{14pt}{11pt}	 \\
		& Our2(40\%) & $\times$  &{76.15}		&{73.60}		&{92.87}	  & {91.61} &{2.55}	 & {1.26} & 53.5  	 \\
		
		& ThiNet\cite{Luo2017} & \checkmark  &{72.88}		&{72.04}		&{91.14}	  & {90.67} &{0.84}	 & {0.47} & 36.7  	 \\
		& AutoPruner~\cite{Luo2020} & \checkmark  &{76.15}		&{74.76}		&{92.87}	  & {92.15} &{1.39}	 & {0.72} & 51.2  	 \\
		
		& SFP(30\%)~\cite{He2018} & \checkmark  &{76.15}		&{62.14}		&{92.87}	  & {84.60} &{14.01}	 & {8.27} & 41.8  	 \\
		
		& ASFP(30\%)~\cite{He2019} & \checkmark  &{76.15}		&{75.53}		&{92.87}	  & {92.73} &{0.62}	 & {0.14} & 41.8  	 \\
		
		& Our1(30\%) & \checkmark  &{76.15}		&{75.98}		&{92.87}	  & {92.81} &{0.17}	 & {0.06} & 41.8  	 \\
		& Our2(30\%) & \checkmark  &{76.15}		&\textbf{76.00}		&{92.87}	  & \textbf{92.90} &\textbf{0.15}	 & \textbf{-0.03} & 41.8  	 \\
		\bottomrule 
	\end{tabular}
\end{table*}


\begin{table}[ht]\small
	\vspace{-2mm}
	\renewcommand{\arraystretch}{1.3}
	\caption{ COMPARISON OF THE LINEAR DECAY STRATEGY AND EXPONENTIAL DECAY OF SRFP ON CIFAR-10
	}
	\vspace{-1mm}
	\label{table:decay_strategy}
	\vspace{-4mm}
	\setlength{\tabcolsep}{0.2em}
	\begin{center}
		\begin{tabular}{ c  c  c  c   c } \toprule
			\multirow{2}{*}{Model} & \multirow{2}{*}{\shortstack{Pruned\\percent(\%)}}& \multirow{2}{*}{\shortstack {SFP\\Accu.(\%) }} & \multirow{2}{*}{\shortstack {Linear\\Accu.(\%) }} & \multirow{2}{*}{\shortstack {Exp. \\Accu.(\%)}} \\
			&        &         &          &                  \\ 
			\midrule
			ResNet-56      & 30   & 93.05 $\pm$ 0.17  &  93.03 $\pm$ 0.37   & \textbf{93.22} $\pm$ 0.38      \\  
			ResNet-56      & 40   & 92.06 $\pm$ 0.63  &   \textbf{92.77} $\pm $ 0.16   & 92.67 $\pm$ 0.41    \\ 
			ResNet-110     & 20  &93.86 $\pm$ 0.45 & \textbf{93.97} $\pm$ 0.53  & 93.61 $\pm$ 0.56   \\ 
			\bottomrule
		\end{tabular}
	\end{center}
	\vspace{-2mm}
\end{table}


\textbf{Settings}. On ILSVRC-2012, owing to the excellent performance of our ASRFP method on CIFAR-10, we mainly focus on the evaluation of our ASRFP on ResNet-18/34/50, especially comparing with that of ASFP. By default, we use exponential decay with $\epsilon=1e-7$ and $\alpha_0=1$. For pre-trained models, we let $\epsilon=1e-9$.

\textbf{Results.} 
We summarize the results of both SRFP and ASRFP in Table~\ref{table:imagenet_accuracy}. Here, we use "Our1" and "Our2" to refer to SRFP and ASRFP individually. According to Table~\ref{table:imagenet_accuracy}, our ASRFP still outperforms other pruning methods in most cases, especially for networks with large pruning rates like 40\%. It is worth noting that the relative advantage of our ASRFP method is larger in Top-1 accuracy than in Top-5 accuracy.


\textbf{Convergence Analysis}. To illustrate the intrinsic mechanism of SRFP and ASRFP, we compare different Test Accuracy Drops of ResNet-34 on ILSVRC-2012 among SFP/ASFP/SRFP/ASRFP with the training epochs increasing when the pruning rate is 30\%, as shown in Figure~\ref{fig:acc_gap}. The Test Accuracy Drop is the difference between the Top-1 accuracy before pruning and the Top-1 accuracy after pruning, where 0 means that there is no obvious accuracy drops caused by pruning.

SRFP, ASRFP and ASFP are all variants of SFP, pursuing better performance at the cost of slowing down the speed of convergence. We notice that the Test Accuracy Drop of SFP converges to 0 at the fastest speed and that of ASRFP converges to 0 at the slowest speed because ASRFP softens the SFP in both pruning rates and the weight decay of pruned filters, thus requiring more epochs to converge, while ASFP and SRFP soften the SFP in pruning rates and the weight decay of pruned filters respectively.


\subsection{Ablation Study}\label{1-norm and 2-norm}
We conducted a series of ablation experiments.

\begin{figure}[t]
	\center
	\vspace{-4mm}
	\subfigure[ResNet-56]{
		\label{fig:diff_alpha0_56}
		\includegraphics[width=0.49\linewidth]{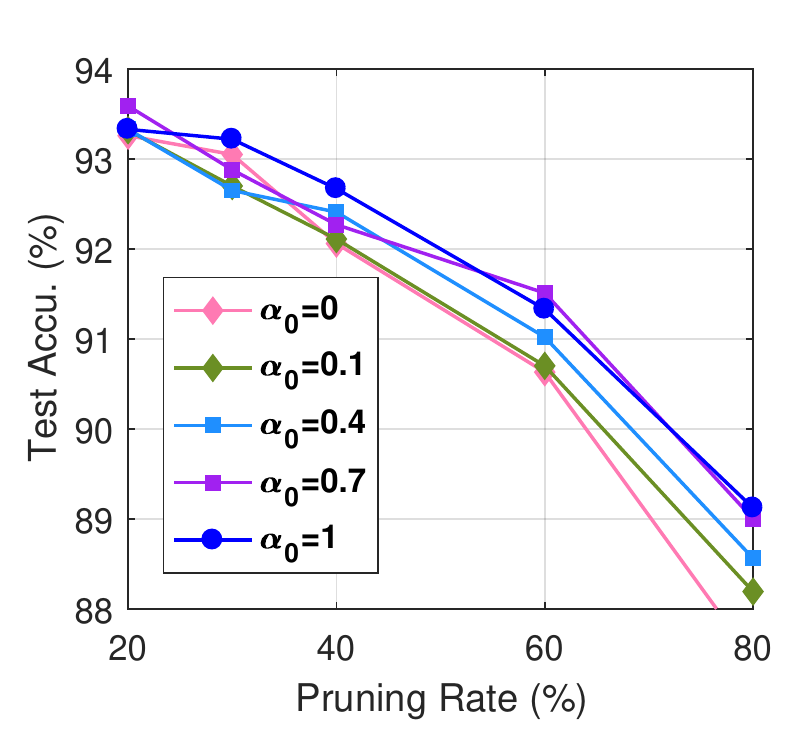}
	}
	\hspace{-6mm}
	\subfigure[ResNet-110]{
		\label{fig:diff_alpha0_110}
		\includegraphics[width=0.49\linewidth]{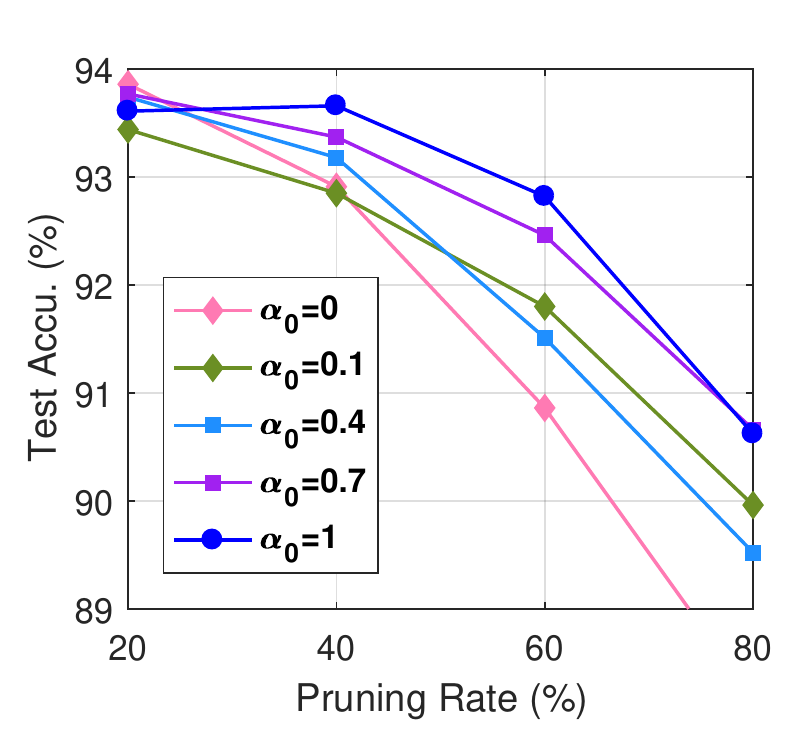}
	}
	\vspace{-2mm}
	\caption{
		Comparison of test accuracies of different pruning rates and $\alpha_0$ for ResNet-56/110 on CIFAR-10 between SFP ($\alpha_0=0$) and our SRFP ($\alpha_0 \neq 0$) with the linear decay strategy. SFP is a special case of SRFP with $\alpha_0=0$.
	}
	\vspace{-3mm}
	\label{fig:resnet_cifar100}
\end{figure}

\textbf{Varying pruning rates and $\bm{\alpha_0}$.}
To further shed light on the performance of the SRFP method, we present the results of the estimation of the accuracy of diverse pruning rates and $\alpha_0$ for ResNet-56/110 in Figure~\ref{fig:diff_alpha0_56} and Figure~\ref{fig:diff_alpha0_110}.
Note that SFP is a special case of SRFP with $\alpha_0=0$. Evidently, $\alpha_0=1$ is a remarkable choice across different pruning rates and network architectures. Besides, as the pruning rate increases, the relative advantage of SRFP with $\alpha_0 > 0$ is enlarged compared with SFP.

\begin{figure}[!t]
	\vspace{-1mm}
	\begin{centering}
		\includegraphics[width=0.28\textwidth]{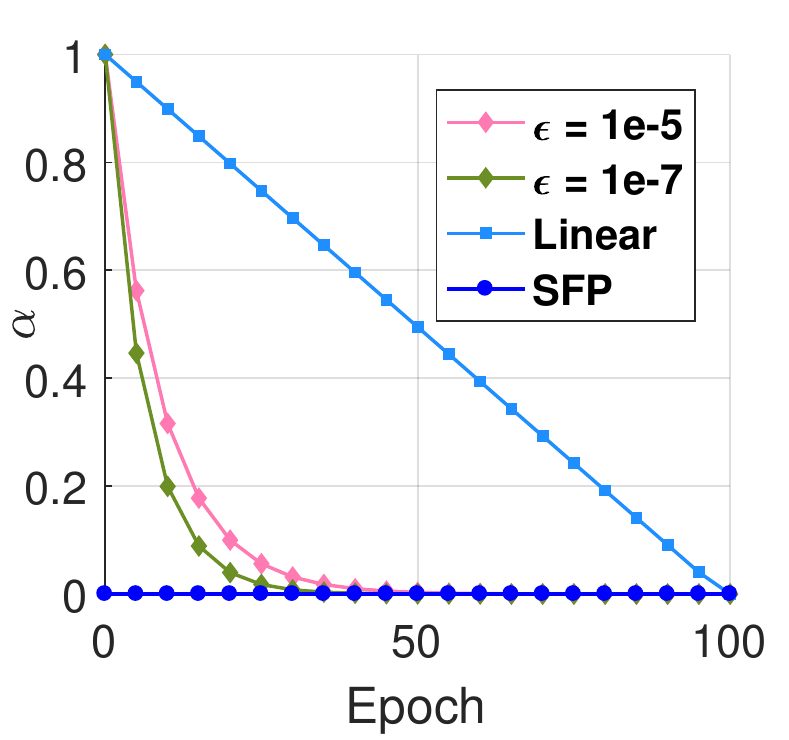} 
		\par\end{centering}
	\vspace{-2mm}
	\caption{
		Comparison of linear decay and exponential decay with various $\epsilon$ to control the speed of weight decay. All strategies except SFP use $\alpha_0=1$.
	}
	\vspace{-3.7mm}
	\label{fig:decay_strategy} 
\end{figure}

\textbf{ Different decay strategies.}
We compare the results of linear decay and exponential decay on CIFAR-10, as shown in Table~\ref{table:decay_strategy}.  Both strategies set $\alpha_0=1$. SFP is a special case of SRFP when $\alpha_0=0$.  We use exponential decay with $\epsilon=1e-5$ for CIFAR-10 and $\epsilon=1e-7$ for ILSVRC-2012 by default. Although linear decay is simple compared with exponential decay and performs well on CIFAR-10, the linear decay strategy produces very poor results on ILSVRC-2012, due to the limited training and pruning epochs. As presented by Figure~\ref{fig:decay_strategy}, linear decay strategy decreases the pruned weights in a much smoother and slower manner compared with exponential decay, leading to the disastrous non-convergence issue on complex dataset like ILSVRC-2012. 
And we evaluate different values of $\epsilon$ for ResNet-18 trained from scratch with the pruning rate of 40\% on ILSVRC-2012, including $1e-5$, $1e-7$ and $1e-9$, ultimately choosing $\epsilon=1e-7$, trading off between the accuracy and convergence.

\section{Conclusion}

To conclude, we propose a pruning method SRFP and its variant ASRFP, softening the pruning operation of SFP and ASRFP. We present two kinds of weight decay strategies, exponential decay and linear decay and investigate their differences. Our methods perform well across various networks, datasets and pruning rates, also transferable to weight pruning. In theory, our methods are doing the $\ell_2$-norm regularization on those pruned filters. Besides, we study the intrinsic mechanism of SRFP and ASRFP and note that SRFP, ASRFP and ASFP pursue better results while slowing down the speed of convergence. 

\bibliographystyle{IEEEtran}
\bibliography{softer}

\end{document}